\documentclass[10pt,twocolumn,letterpaper]{article}

\usepackage{cvpr}              

\usepackage{graphicx}
\usepackage{amsmath}
\usepackage{amssymb}
\usepackage{booktabs}
\usepackage{textcomp}
\usepackage{gensymb}
\usepackage{amsmath}
\usepackage{algorithmic}
\graphicspath{{images/}}
\usepackage{multirow}
\usepackage{arydshln}

\usepackage[pagebackref,breaklinks,colorlinks]{hyperref}
\usepackage[capitalize]{cleveref}
\crefname{section}{Sec.}{Secs.}
\Crefname{section}{Section}{Sections}
\Crefname{table}{Table}{Tables}
\crefname{table}{Tab.}{Tabs.}

\def\cvprPaperID{13} 
\def\confName{CVPR-VCAD}
\def\confYear{2023}

\newcommand\blfootnote[1]{%
	\begingroup
	\renewcommand\thefootnote{}\footnote{#1}%
	\addtocounter{footnote}{-1}%
	\endgroup
}

\usepackage[numbers,sort&compress]{natbib}

\usepackage{color}
\definecolor{mmcolored}{rgb}{0.0,0.0,0.0}
\newcommand{\mm}[1]{\textcolor{mmcolored}{#1}}
\newcommand{\pa}{\textcolor{black}}
\newcommand{\rpa}{\textcolor{black}}
\definecolor{mmcolored2}{rgb}{0.0,0.0,0.0}
\newcommand{\mmm}[1]{\textcolor{mmcolored2}{#1}}

\begin{document}
	
	\title{LeTFuser: Light-weight End-to-end Transformer-Based Sensor Fusion for Autonomous Driving with Multi-Task Learning}
	
	\author{Pedram Agand* ~~~
		Mohammad Mahdavian* ~~~
		Manolis Savva ~~~
		Mo Chen\\
		Department of Computing Science, Simon Fraser University, BC, Canada.\\
		{\tt\small \{pedram\_agand, mohammad\_mahdavian, manolis\_savva, mochen\}@sfu.ca}
	}
	\maketitle
	\blfootnote{* Equal contribution}
	\begin{abstract}
		In end-to-end autonomous driving, the utilization of existing sensor fusion techniques \mmm{and navigational control methods} for imitation learning proves inadequate in challenging situations that involve numerous dynamic agents. To address this issue, we introduce LeTFuser, a \mmm{lightweight} transformer-based algorithm for fusing multiple RGB-D camera representations. To perform perception and control tasks simultaneously, we utilize multi-task learning. Our model comprises of two modules, the first being the perception module that is responsible for encoding the observation data obtained from the RGB-D cameras. Our approach employs the Convolutional vision Transformer (CvT) \cite{wu2021cvt} to better extract and fuse features from multiple RGB cameras due to local and global feature extraction capability of convolution and transformer modules, respectively. 
		
		\mmm{Encoded features combined with static and dynamic environments are later employed by our control module to predict waypoints and vehicular controls (e.g. steering, throttle, and brake).}
		We use two methods to generate the vehicular controls levels. The first method uses a PID algorithm to follow the waypoints on the fly, whereas the second one directly predicts the control policy using the measurement features and environmental state. We evaluate the model and conduct a comparative analysis with recent models on the CARLA simulator using various scenarios, ranging from normal to adversarial conditions, to simulate real-world scenarios. \mmm{Our method demonstrated better or comparable results with respect to our baselines in term of driving abilities.}
		The code is available at \url{https://github.com/pagand/e2etransfuser/tree/cvpr-w} to facilitate future studies.
	\end{abstract}
	
	\section{Introduction}
	\label{sec:intro}
	\mm{Many works in the autonomous driving literature \mmm{have} been} focusing on different aspects of perception and control tasks for safe navigation   \cite{chen2020learning,ohn2020learning,behl2020label,prakash2020exploring,zhao2021sam,toromanoff2020end,velasco2020autonomous}. 
	Recent advances in end-to-end driving \mm{neural network (NN)  models} have demonstrated remarkable results using single modality inputs, such as \mm{image}   
	and \mm{LiDAR}  \cite{filos2020can}. 
	However, these approaches face limitations in complex urban scenarios involving adversarial situations due to their lack of 3D scene understanding  \cite{dosovitskiy2017carla}. Sensor fusion has shown promise in addressing these challenges by integrating multiple sensor modalities, such as cameras and LiDAR sensors, to create a more comprehensive scene representation \cite{feng2020deep,agand2022human}.
	~Despite the improvements, these fusion methods often require large computational resources and face challenges in balancing learning signals between perception and control tasks  \cite{xiao2020multimodal}. 
	Moreover, integrating multiple modalities with different data shapes and representations requires sophisticated preprocessing techniques \pa{such as ELPP  \cite{d2016earlinet}, SaDMS \cite{liu2017efficient}}, leading to increased model complexity and the potential for information loss.
	
	Recent advancements in end-to-end autonomous driving have explored the integration of different sensor modalities, such as LiDAR, and RGB-D cameras, to enhance performance. \mmm{\citet{xiao2020multimodal} employs} \mmm{a} Convolutional Neural Network (CNN) to extract data features provided by \mmm{an} RGB-D camera and \mmm{produce} future vehicle waypoints and navigational signals. Another well-known method is TransFuser \cite{Chitta2022PAMI}, which uses a multi-modal fusion transformer to incorporate global context and pairwise interactions into the feature extraction layers of different input modalities.
	\rpa{We were inspired by these methods and demonstrated that combining ideas from these approaches could potentially lead to more robust and accurate end-to-end autonomous driving solutions. \mmm{On the perception side,} we leverage the strengths of both global and local context reasoning provided by transformers and CNNs. \mmm{In the control side, we take advantage of trajectory-guided control, which we introduce next.}}
	
	\mmm{The Trajectory-guided Control Prediction (TCP) \cite{wu2022trajectory} framework, which combines trajectory planning and direct control is a multi-task learning framework for end-to-end autonomous driving.
		By incorporating a multi-step control prediction branch with a dynamic branch and trajectory-guided attention, TCP can \mmm{improve} temporal reasoning and achieve superior performance in the CARLA driving simulator, even surpassing methods using multiple cameras and LiDAR sensors.}
	
\begin{figure*}[t]
		\centering
		\centerline{\includegraphics[width=1\linewidth,trim={0cm 0cm 9cm 0cm },clip]{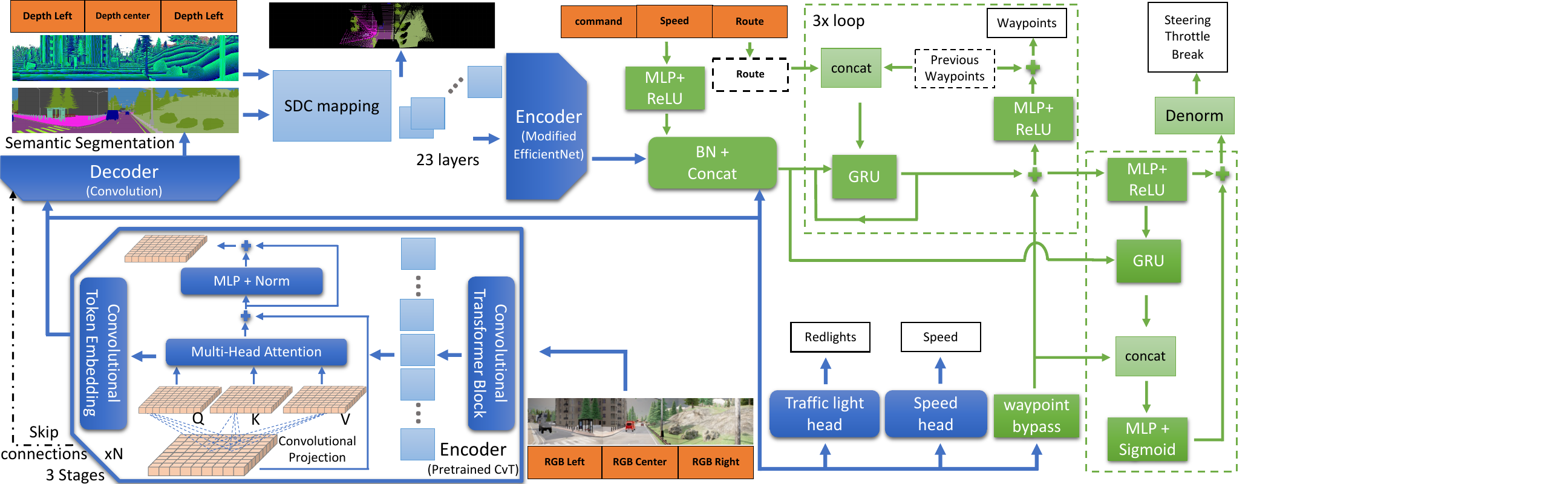}}
		\caption{Model architecture: It consists of trainable and non-trainable components represented by light and dark-colored items, respectively. The blue-colored items \mmm{represent the perception module}, while the controller module \mmm{and inputs} are represented by green and orange\mmm{, respectively.} \rpa{The white boxes represent different tasks that are learned simultaneously}.  The process inside the \rpa{top dashed green box} is iterated three times during the training, in which the model predicts waypoints \mmm{to} estimate the \mmm{values} of vehicular controls independently.}
		\label{fig:structure}
	\end{figure*}

	In this paper, \mmm{inspired by these three methods,} we propose a novel \mm{deep neural network architecture for end-to-end autonomous driving} that leverages the complementary advantages of RGB and depth information provided by an RGB-D camera, addressing the challenges faced by existing single modality and sensor fusion approaches \mmm{as well as navigational commands prediction}. Our model consists of two main modules \mmm{shown in Fig.~\ref{fig:structure}}: the  \textit{perception module}, which encodes high-dimensional observation data and performs semantic segmentation, semantic depth cloud (SDC) mapping \mm{and ego vehicle speed and traffic light prediction}; and the  \textit{control module}, which decodes the features encoded by the perception module along with additional GPS, command and speedometer information to predict waypoints and control policy.
	
	In the perception module, we utilize Convolutional vision Transformers (CvT) \cite{wu2021cvt} and \mm{EfficientNet  \cite{tan2019efficientnet} to adeptly extract RGB image and SDC map features. \pa{We then} fuse them using a CNN-based fusion layer}. Additionally, we employ two agents in the control module to process the perception module's outputs, fostering diversified and resilient decision-making. To tackle the issue of balancing learning signals, \rpa{similar to \citet{xiao2020multimodal}}, we implement a Modified Gradient Normalization (MGN) method, ensuring uniform learning pace across all tasks.
	Finally, we evaluated our model on the CARLA simulator with various scenarios including normal-adversarial situations, \mmm{reported in Section~\ref{results}} demonstrating improved performance over baseline methods. \mmm{We achieve better and comparable results with respect to our baselines in short and long paths, respectively.}
	
	
	\section{Related works}
	\label{sec:rw}
	\mmm{\textbf{Multi-Modality}}: Recent advancements in multi-modal end-to-end \mm{autonomous} driving have highlighted the potential of using RGB images alongside depth and semantic information to enhance driving performance \cite{zhou2019does}. \mmm{Recently, a few } studies \rpa{by~\citet{xiao2020multimodal}} and \mmm{\citet{behl2020label} } have investigated the effectiveness of incorporating depth and semantic data as intermediate representations for driving tasks. In our work, we focus on combining \pa{RGB} and \mm{Depth} inputs, \mmm{which are readily available in autonomous vehicles and provide complementary} scene representations. 
	
	\mmm{\textbf{Sensor Fusion}}: Most sensor fusion research has focused on perception tasks such as object detection  \cite{fadadu2022multi,chen2021roifusion,liang2019multi} 
	and motion forecasting \cite{zhang2020stinet,djuric2021multixnet,meyer2020laserflow}. 
	These approaches typically \mm{include} multi-view LiDAR \mm{data} or combine camera input with LiDAR data by projecting features between different spaces. ContFuse \cite{liang2018deep_old} is an approach fusing multi-scale \pa{RGB} and LiDAR bird's eye view (BEV) features densely. However, these methods do not capture the global context of the 3D scene, which is crucial for safe navigation in challenging scenarios. 
	\mmm{In this work, we use CNNs to fuse the multi-modal data received by the RGB-D sensors.}
	
	\mmm{\textbf{Bird's Eye View Strategies}}: In this domain, researchers \mm{either use LiDAR data or} fuse RGB images and depth maps from a single RGB-D camera \cite{natan2022end}. They project the depth map \mm{with the semantic segmentation to} create a semantic depth cloud (SDC) \mmm{from a BEV angle}. \mmm{This way the model benefits from the clearer delineation of} occupied or \mm{navigable} regions provided by the SDC's semantic information compared to LiDAR point clouds containing only height data \cite{prakash2021multi, filos2020can}. \rpa{\citet{huang2020multi}} fused RGB images and depth maps to capture a deeper global context, while \rpa{\cite{prakash2021multi}} combined RGB images and preprocessed LiDAR point clouds to leverage different perspectives, such as front-view and BEV. These approaches used either high-level navigational commands \rpa{as described by \citet{huang2020multi}} or sparse GPS locations provided by a global planner \rpa{as explored by \citet{prakash2021multi}} for driving. In our research, we consider using \mmm{waypoints} instead of high-level navigational commands as \mmm{waypoints are more informative and} better reflects real-world autonomous driving conditions \cite{guo2017toward_old}. 
	
	\mmm{\textbf{Imitation Learning}}: Studies in end-to-end autonomous driving usually fall into two categories: reinforcement learning (RL) and imitation learning (IL). \rpa{\citet{liang2018cirl_old, kendall2019learning}} have shown the potential of RL, while IL approaches such as LBC \cite{chen2020learning} and NEAT \cite{chitta2021neat} have demonstrated impressive performance. Our work adapts the auto-regression scheme used in TransFuser and its variants \cite{prakash2021multi, jaeger2021expert,agand2023online}.
	
	\mmm{\textbf{End-to-End Autonomous Driving}}: End-to-end multi-task learning approaches offer benefits in training efficiency and integration simplicity. Imitation learning-based methods have been investigated  for autonomous driving tasks, \mmm{by} exploring additional perception tasks to improve feature extraction \cite{chen2020learning}. Combinations of various autonomous driving \mmm{subtasks}, such as object detection, lane detection, semantic segmentation, and depth estimation have been proven to achieve incredible performance \cite{wu2022yolop,chen2022persformer}. 
	In our work, we adopt a similar multi-task learning approach, but utilize depth from an RGB-D camera as input \cite{feng2020deep}. 
	We address the imbalanced learning problem in multi-task learning by implementing a MGN algorithm  \cite{natan2022towards}.
	
	\section{Methodology}
	In this section, we present the details of our proposed approach. Our model is structured around two key components: the perception and control modules, each fulfilling a specific set of tasks to enable vehicle navigation. \mm{In the perception \mmm{module}, we extract features from RGB and a BEV SDC map.} Also, the model \mmm{attempts to }accurately predict traffic lights and ego vehicle speed from the RGB embedded features. In the subsequent control \mmm{module}, our model utilizes the extracted features, current ego vehicle speed, and GPS location to provide reliable waypoints and vehicular commands.
	
	\subsection{Perception Module}
	\pa{As illustrated in Fig.~\ref{fig:structure}, the perception module serves to extract features from RGB images using CvT \cite{wu2021cvt} to perform semantic segmentation and \mmm{later} generate a BEV SDC map}. Our perception module receives a total of three RGB images from three vehicle cameras, with the first camera capturing the front view angle and the other two tilted to the left and right by 60 degrees. To provide a comprehensive understanding of the environment, depth images are also captured from each camera. The front RGB and depth images have a \mm{resolution} of \mm{160 $\times$ 320}, while the non-overlapping side cameras capture images with the \mm{resolution of 160 $\times$ 224. This results in a total of 160 $\times$ 768} pixels for both RGB and depth images. 
	
	\subsubsection{Convolutional Vision Transformer}
	\mm{The cornerstone of our feature extractor is the CvT} \cite{wu2021cvt} that has been pretrained on the ImageNet \cite{deng2009imagenet} \mm{shown in bottom left section of the Fig.~\ref{fig:structure}}. \mmm{This module is responsible for extracting features from the three RGB images.} This particular \mm{network} was selected for its unique ability to leverage both convolution and transformer modules, \mmm{greatly facilitating feature extraction}. Convolution \mm{layers} excel at extracting local features, while transformers are known for their \mm{ability} in global feature extraction and learning. By combining the strengths of these two powerful techniques, CvT ensures that our feature extractor is capable of capturing both local and global features, resulting in \mmm{better} visual representations. 
	
	\mm{CvT-13 \cite{wu2021cvt}}, a light version of the CvT, has been carefully selected for its exceptional performance and fewer number of parameters. As one can see in Fig.~\ref{fig:structure}, it has been designed with three main stages, each of which incorporates a Convolutional Token Embedding (CTE) to process the 2D input \mm{images}. The features extracted by the CTE are then normalized and passed through a Convolutional Transformer Block (CTB). \mm{To apply both convolutional and attention layers,} the CTB uses a depth-wise separable convolution operation known as Convolutional Projection to create the query, key, and value embeddings. These embeddings are then passed through a transformer module to extract global features, ultimately resulting in highly accurate and comprehensive feature maps. The last layer of CvT is a fully connected layer used for image classification that we remove from the model. As a result, the RGB extracted features contains 384 features, each with a size of $10\times48$.
	
	\subsubsection{Semantic Segmentation}
	After the feature maps are extracted, \mm{we use them in different sections of the model, as they contain valuable information. First,} they are utilized to train a semantic segmentation decoder capable of accurately identifying 23 different classes \mm{depicted in the Fig.~\ref{fig:structure} as ``Decoder''. Later, we use this to create a BEV SDC map}. To achieve this, we have developed a segmentation decoder that consists of three convolutional layers and a final pointwise convolution with sigmoid activation.
	By leveraging skip connections, we can effectively capture both local and global features, resulting in accurate and comprehensive segmentation maps. 
	
	\begin{figure*}[t]
		\centering
		\centerline{\includegraphics[width=1\linewidth]{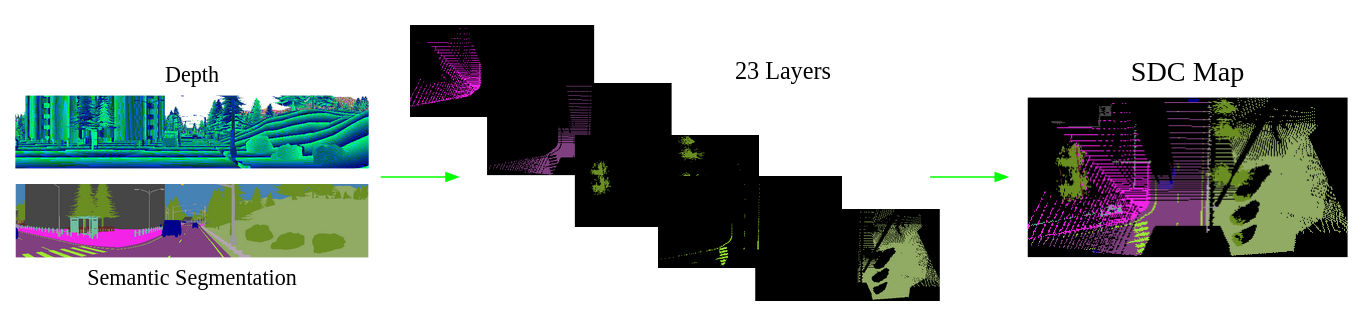}}
		\caption{Semantic Depth Cloud (SDC) map created from three depth and semantic segmentations acquired from the three cameras attached to the vehicle. The SDC mapping creates three separate BEV maps using estimated semantic segmentation and attaches them together.}
		\label{fig:sdc_map}
	\end{figure*}
	
	\subsubsection{Semantic Depth Cloud}
	In order to enhance the understanding of the scene, we have utilized a method that involves the creation of a BEV \mm{Semantic Depth Cloud} using the provided depth images and estimated semantic segmentation, in addition to the RGB images \cite{natan2022end}. \mm{The SDC map represents the ego-vehicle's surrounding environment and contains richer information with respect to the LiDAR data due to the \rpa{23 semantic class layers}. Each layer represents one class of environment object.} \mmm{As shown in} the Fig.~\ref{fig:sdc_map}, the SDC is created without considering the height information \mmm{($y$-axis)}, as a BEV perspective requires only the \mm{$x$-axis and $z$-axis}. The process of creating the SDC involves separately generating a SDC map for each camera, and then merging them together after rotating the side maps with an appropriate angle. 
	
	We define a 64-meter distance range in front of the vehicle and 32 meters to each side of the camera, creating a coverage area of $64 \times 64$ square meters. The SDC maps have dimensions of $160 \times 320$ square centimeters for the front and $160 \times 224$ for the sides. We generate a transformation matrix for the \mmm{$x$-axis} using camera parameters and normalize the coordinates to align with the SDC tensor spatial dimension. One-hot encoding is applied to yield a 23-channel SDC tensor. The resulting maps are copied to an empty tensor with dimensions of $160 \times 768$, with side maps rotated at a 42-degree angle. For feature extraction, we use the compact EfficientNet-B1 \cite{tan2019efficientnet} network to generate a tensor of 192 features, each with a size of $10\times48$.

	\subsection{Controller Module}
	
	The control module, depicted as green squares in the Fig. \ref{fig:structure}, receives \mm{RGB, SDC and navigational measurements} features and uses them to predict the vehicle's future waypoints. \pa{The navigational measurements includes route location, navigational command provided by the global planner and the ego vehicle speed. The navigational command specifies the vehicle's general direction, such as left, right, forward, stop, etc., and is defined as a one-hot vector.} Subsequently, \mm{the control module} predicts the appropriate vehicular control, including steering, throttle, and brake, based on the predicted waypoints and fused features.  To predict the vehicle's future waypoints, we employ a gated recurrent unit (GRU) similar to \rpa{\citet{natan2022end}}. The GRU is a suitable choice as it addresses the vanishing gradient problem while maintaining a better performance-cost ratio compared to other RNN methods. \pa{To train a control model that predicts current control actions based on current input, behavior cloning is commonly used but relies on the assumption of independent and identically distributed (IID) data, which is not valid for closed-loop tests \cite{wu2022trajectory}. To overcome this issue without reinforcement learning, we used a similar trick to \rpa{\citet{wu2022trajectory}} for predicting multi-step control actions into the future. To this end, we first employ a waypoint branch that utilizes fused features and environment-agent static knowledge through waypoint bypass. \rpa{Further, we deploy a dynamic branch to capture the environment-agent dynamic interaction given the learned static knowledge. The dynamic branch provides dynamic information such as object motion and traffic light changes, while the waypoint branch incorporates static information like curbs and lanes and improves spatial consistency across both branches.}}
	
	\mm{In order to fuse the extracted features from RGB images and the SDC map}, \pa{we concatenate them and apply batch normalization (BN) to them and then concatenate \mmm{the result} with the measurement \mm{tensor}.}  The GRU in the waypoint branch  takes the fused features as the initial hidden state, and the inputs include the current waypoint in the BEV space, the route location coordinate transformed to the BEV space.  The initial waypoint coordinate is always positioned at $(0, 0)$, the bottom-center point of the SDC map. To transform the global coordinates $(x_g, y_g)$ to local coordinates $(x_l, y_l)$, we use the Eq.~\ref{global_to_local}. 
	We then add the next hidden state from the GRU to waypoint bypass, which is the RGB features that have passed through a biasing module. The biasing module consists of adaptive global pooling and a linear layer applied to the RGB extracted features. Finally, a sigmoid function is applied to convert all values between 0 and 1. We apply a multi-layer perceptron (MLP) network containing two linear layers and a rectified linear unit (ReLU) to the biased GRU hidden state to obtain normalized control commands in the range of 0 to 1.
	\begin{equation} \label{global_to_local}
		\begin{bmatrix}
			x_l\\
			y_l
		\end{bmatrix}
		= 
		\begin{bmatrix}
			cos(90\degree + \theta_v) &  -sin(90\degree + \theta_v)\\
			sin(90\degree + \theta_v) &~ cos(90\degree + \theta_v)
		\end{bmatrix}^{\top}
		\begin{bmatrix}
			x_g - x_vg\\
			y_g - y_vg
		\end{bmatrix}
	\end{equation}
	
	\pa{The GRU in the dynamic branch takes the same fused features as the waypoint branch for the initial hidden state to improve consistency and the inputs include the predicted vehicular command from the waypoint branch. The result \mmm{is then concatenated} with the same waypoint bypass, representing the abstract static coarse simulator and then fed to MLP and sigmoid to create adjusted control output. }
	
	To determine the suitable vehicular control, we compute them in two different ways. First, we denormalized the summation of the predicted vehicular command from waypoint branch and adjusted control from dynamic branch. Second, we use two separate PID controllers to predict the vehicle controls, one for finding the steering command (lateral) and the other for finding the throttle and brake (longitudinal), using the predicted waypoints and the current speed. Our control policy is similar to \rpa{\citet{natan2022end}} that calculates the control commands using both methods based on the scenario.
	During driving, the vehicle relies on the first two waypoints to calculate its next destination by taking their average. However, we also generate a prediction for a third waypoint, which supplies additional data to the GRU and waypoint prediction layer. Furthermore, this approach enables the MLP and PID agents to receive identical information, since the last biased hidden state contains the details from the second waypoint.
	
	\section{Experiments} \label{experiments}
	\label{sec:ex} 
	
	\subsection{Dataset}
	We use CARLA \cite{dosovitskiy2017carla} (0.9.10) \pa{for the \mmm{simulation} environment  which has} 8 available towns for training and testing. \pa{We \mm{train our model on the} 210 GB  publicly available dataset \mmm{presented in the} TransFuser \cite{Chitta2022PAMI} for the experiment.} All 8 towns \pa{were} used for training and the dataset includes approximately 2500 routes through junctions with an average length of 100m and \mmm{approximately} 1000 routes \mmm{including} curved highways with an average length of 400m. To generate data for training purposes, an expert policy is formulated, which employs privileged information obtained from the simulator to control the driving process \rpa{inspired from \citet{chen2020learning}}. The expert's waypoints serve as ground-truth labels for the imitation loss, making \pa{the expert} comparable to an automatic labeling algorithm. 
	\rpa{To accomplish lateral control, the expert policy follows the path generated by the A* planner, and a PID controller is used to minimize the angle of the vehicle towards the next waypoint in the route, which is at least 3.5 meters away. Meanwhile, longitudinal control is performed using a version of model predictive control, which differentiates between 3 target speeds.} The standard target speed is 4.0 m/s, but the speed is reduced to 3.0 m/s when the expert is inside an intersection. Additionally, if an infraction is predicted, the target speed \mm{changes} to 0.0 m/s, bringing the vehicle to a halt. 
	The longitudinal and lateral controllers use PID values of $K_p = 5.0, K_i = 0.5, K_d = 1.0$ and $K_p = 1.25, K_i = 0.75, K_d = 0.3$, respectively. For the running average of both controllers' integral term, we use a buffer of size 40.
	
	Originally, \mm{all three} RGB images and depth maps are retrieved at a resolution of \mm{$480 \times 960 $ then cropped to $160 \times 320$} to avoid distortion. Thus, \mm{all three} RGB images are represented as $R \in {0, \ldots, 255}^{3\times160\times320} $ 
	\mmm{attached to each other side-by-side to create a single image. \mmm{The same process is applied to the depth images to create one single depth image.} In order to calculate the \mmm{real} depth value for each pixel i in the depth map we use the following}:
	\begin{equation}
		{R}_i^{dec} = \frac{\mmm{256^2B_i+256G_i+R_i}}{256^3-1}\times 1000,
	\end{equation}
	
	\noindent where \rpa{$(R_i, G_i, B_i)$ are the stored values of pixel i. Also, since the stored values are 8-bit, they have a maximum of 255, and 1000 here is the \mmm{maximum} depth range of the RGB-D camera in meters. All together they result in ${R}_i^{dec}$ which represents the true depth of pixel i.} 
	\mmm{Also, each semantic segmentation ground truth is represented as $S \in
		{0, 1}^{23\times160\times320}$. It contains 23 classes of data that contain value of 1 or 0 based on whether the pixel belongs to that class or not}. The object classes for the semantic segmentation are according to \rpa{\citet{natan2022end}}.
	Moreover, the waypoints are represented in BEV space with ${\omega\rho_i = (x_i, y_i)}^3_{i=1}$. \mmm{The center of the BEV space, marked by coordinates $(0,0)$ in the local vehicle coordinate system, is positioned on the ego vehicle itself, at the bottom-center point. The model assesses vehicular controls within a normalized range spanning from 0 to 1, subsequently denormalizing them to their original values, encompassing steering within the range of $[-1,1]$, throttle ranging from $[0,0.75]$, and brake as either 0 or 1.}
	
	\mmm{The traffic light state value changes to 1 if a red light appeared, otherwise it is stated as 0. We encapsulate the speed measurement (in m/s) and GPS locations into a one-hot encoded vector as a high level navigational command.}
	
	\mmm{\subsection{Task and Scenario} \label{task_scenario}
		The task entails navigating through a variety of areas such as highways, cityscapes, and residential neighborhoods along the predefined paths consisting of sparse goal locations specified in the GPS coordinates provided by a global planner. The routes encompass numerous scenarios, each initialized at predefined positions, designed to evaluate the agent's capability to adapt wide spectrum of adversarial situations and varying weather conditions.}
	
	The first scenario, called 1WN, involves training the model on all available maps and route sets except for the Town05 , which are reserved for validation. \pa{The model is then evaluated on both Town05 short and long routes, consisting of 32 short and 10 long routes, to assess its performance}. The evaluation is conducted in clear noon condition in a normal situation, and all non-player characters follow the traffic rules.  In second scenario, 1WA, the model is tested under \mmm{adversarial} non-player characters (NPC) behavior, which may lead to collisions, such as pedestrians suddenly crossing the street or bicyclists appearing. Additionally, the traffic light manager may intentionally create a state with double green lights at an intersection, simulating emergency situations where an ambulance or firefighter may skip the traffic light. Along with properly driving the ego vehicle, the model is expected to react safely \mmm{to the environment changes and noises and avoid accidents} in these adversarial situations.
	
	\subsection{Implementation Details}
	\mmm{To effectively gather knowledge across multiple tasks using end-to-end learning, we predefined a set of distinct loss functions. The comprehensive loss that encompasses all tasks can be calculated as follows}:
	\begin{equation}
		\begin{split}
			\mathcal L_\text{TOTAL}=\alpha_1\mathcal{L}_\text{SEG}+\alpha_2\mathcal{L}_\text{ST}+\alpha_3\mathcal{L}_\text{TH}+ \\ \alpha_4\mathcal{L}_\text{BR}+\alpha_5\mathcal{L}_\text{WP}+\alpha_6\mathcal{L}_\text{TL}+\alpha_7\mathcal{L}_\text{SS}+\alpha_8\mathcal{L}_\text{VE}
		\end{split}
	\end{equation}
	
	\noindent where steering loss ($\mathcal{L}_\text{ST}$), throttle loss ($\mathcal{L}_\text{TH}$), brake loss ($\mathcal{L}_\text{BR}$), waypoints loss ($\mathcal{L}_\text{WP}$), traffic light state loss ($\mathcal{L}_\text{TL}$), stop sign loss ($\mathcal{L}_\text{SS}$), and velocity loss ($\mathcal{L}_\text{VE}$) are all simple L1 loss functions. Also, the loss weight for each task is denoted by $\alpha_1, \alpha_2, \ldots, \alpha_8$. \mmm{For the purpose of calculating the semantic segmentation loss function ($\mathcal{L}_\text{SEG}$), a mixture of binary cross-entropy and dice loss is utilized and can be computed through following equation.}
	\begin{equation}
		\begin{split}
			\mathcal{L}_{SEG} = \Big (\frac{1}{N}\sum_{i=1}^Ny_i\log(\hat{y}_i)+ &(1-y_i)\log(1-\hat{y}_i)\Big )\\ &+ \Big (1-\frac{2|\hat{y}\cup y|}{|\hat{y}|+|y|}  \Big )
		\end{split}
	\end{equation}
	
	\mmm{Here, $N$ represents the number of pixels at the final layer of our semantic segmentation decoder. Subsequently, $y_i$ and $\hat{y}_i$ correspond to the values of the $i$-th element in the ground truth vector $y$ and the prediction vector $\hat{y}$, respectively.}
	
	This approach allows us to simultaneously leverage both distribution-based and region-based loss components, as described \rpa{ by \citet{natan2022towards}}. Emphasizing the significance of enhancing the semantic segmentation task with supplementary loss criteria is imperative, given that the structural integrity of the entire network relies on it. For the case of three predicted waypoints, only ($\mathcal{L}_\text{WP}$) requires averaging.
	
	
	\mmm{In order to adaptively adjust the loss weights for each training epoch, We utilize the MGN algorithm \cite{natan2022towards}}. To achieve this, we employ the Adam optimizer with a decoupled weight decay of 0.001, and train the model until it reaches convergence \cite{loshchilov2017decoupled}. Initially, the learning rate is set to 0.0001 and gradually halved if the validation metric shows no decline for three consecutive epochs. Furthermore, to prevent unnecessary computational expenses, training is halted if there is no progress for 15 consecutive epochs \mm{or reached the maximum of 40 epochs}. Our model is implemented using the PyTorch framework \cite{paszke2019pytorch} and trained on an NVIDIA GeForce RTX-3090 with a batch size of 20.
	
	\begin{table*}[t]
		\caption{\rpa{Performance comparison of LetFuser (ours) with  the baselines: E2E-F/A \cite{natan2022end}, TF-F/A \cite{Prakash2021CVPR}, and Expert}}
		\begin{center}
			\begin{tabular}{cccp{1.2cm}p{1.2cm}p{1.2cm}p{1.2cm}p{1.2cm}p{1.2cm}} 
				\hline
				Experiment &Model & Inputs & \multicolumn{3}{c}{Normal Clear Noon (1WN)} & \multicolumn{3}{c}{Adversarial Clear Noon (1WA)} \\
				\hline
				Town5 &  & &$\uparrow$ DS & $\uparrow$ RC & $\uparrow$ \mmm{IP} & $\uparrow$ DS & $\uparrow$ RC & $\uparrow$ \mmm{IP} \\ 
				\hline
				\multirow{7}{*}{ Short}& E2E-F & 1 RGB-D & 32.814 & 68.284 & 0.468  & 28.359 & 58.792 & 0.480 \\
				&E2E-A  & 3 RGB-D&\underline{48.833} & \underline{75.824} & 0.588  & \underline{37.263} &\underline{73.986} & 0.466 \\
				&TF-F & 1 RGB-L &17.800 & 19.864 & \textbf{0.942} & 23.641 & 24.373 & \textbf{0.953}\\
				&TF-A  &3 RGB-L &12.494 & 16.315 & \underline{0.886} & 11.349 & 14.675 & \underline{0.843}\\
				&\textbf{Ours} & 3 RGB-D & \textbf{66.012} & \textbf{99.717} & 0.663 & \textbf{51.669} & \textbf{91.918} & 0.574\\\hdashline 
				&Expert & * &99.919 & 99.919 & 1.00 & 79.675 & 95.349 & 0.833\\
				\hline
				\multirow{7}{*}{ Long}& E2E-F & 1 RGB-D & 8.381 & \underline{63.633} & 0.194 & 7.601 & \underline{48.114} & 0.246 \\
				&E2E-A  & 3 RGB-D &7.670 & 48.039 & 0.291  & 11.866 & \textbf{52.424} & 0.456 \\
				&TF-F & 1 RGB-L &\textbf{22.456} & 24.509 & \underline{0.950} & \underline{12.964} & 15.393 & \underline{0.910}\\
				&TF-A  &3 RGB-L &7.111 & 7.351 & \textbf{0.963} & 8.829 & 9.001 & \textbf{0.971}\\
				&\textbf{Ours} & 3 RGB-D &\underline{13.943}& \textbf{63.685} & 0.215 & \textbf{20.584} & 47.186 & 0.493\\\hdashline 
				&Expert & * &60.808 & 100 & 0.608 & 23.344 & 58.872 & 0.619\\
				\hline
			\end{tabular}
		\end{center}
		\label{table1}
	\end{table*}
	

	\begin{table}[t]
		\centering
		\caption{\rpa{Model Specifications}}
		\begin{tabular}{c c c c c} 
			\hline
			Model & $\downarrow$ Total Parameters& $\downarrow$ GPU memory  \\ 
			\hline
			E2E-F  & 20985934 & 2920 MB \\
			E2E-A  & 20985934 & 4958 MB \\
			TF-F  & 66218754 & 3898 MB  \\
			TF-A  & 66401154 & 5015 MB  \\
			Ours  & 31331865 & 3761 MB \\ 
			\hline
		\end{tabular}
		\label{table2}
	\end{table}

	\subsection{Evaluation Metrics}
	
	As per the CARLA leaderboard evaluation setting, we have employed the driving score (DS) as our principal metric. The higher the DS value, the more exemplary the driving ability. The DS can be computed using the following:
	\begin{equation} \label{DS}
		\text{DS} = \frac{1}{N_r} \sum_{i=1}^{N_r}\text{RC}_i \times \text{IP}_i
	\end{equation}
	
	To calculate the driving score (DS) for a given route ($\text{DS}_i$), we use the product of two factors: the percentage of the route that was completed correctly ($\text{RC}_i$) and the corresponding infraction penalty ($\text{IP}_i$). We then compute the average of all $\text{DS}_i$ values over the total number of routes ($N_r$) to obtain the final driving score. To calculate $\text{RC}_i$, we divide the distance driven correctly on the route by the total length of the route. This calculation excludes any incorrect paths taken (e.g., driving on sidewalks).
	To calculate $\text{IP}_i$, we use the following formula:
	
	\begin{equation} \label{DS1}
		\text{IP}_i = \prod_{j}^{M} (p_i^j)^{\text{\#infractions}_j},
	\end{equation}
	where $M$ represents the types of infractions for evaluations, the ideal $\text{IP}_i$ at the beginning of the evaluation is 1.0, and it decreases each time an infraction occurs. \rpa{The final RC and IP scores are calculated by averaging over different routes.} We consider same penalties for different infractions as \rpa{\citet{Chitta2022PAMI}}. 
	
	
	
	\subsection{Baselines}
	We have opted to compare our metrics with some of the cutting-edge techniques in autonomous driving. \pa{The necessary inputs for inference time in each method are presented in the inputs column of the Table~\ref{table1}, where RGB-D denotes the RGB camera and depth information, whereas RGB-L represents the RGB camera and LiDAR information. ``-F" denotes inputs with only the front sensors, while ``-A" signifies inputs of all left, front, and right sensors.}
	
	\pa{As our first baseline, we have selected \mmm{E2E} \cite{natan2022end}, which is an end-to-end approach that uses RGB-D data}.  This algorithm mainly relies on  CNN and EfficientNet to extract features for identifying the \pa{vehicle's} navigational waypoints. \pa{We trained two different versions of \mmm{E2E}: the first one (E2E-F) is identical to the model presented in the paper, and the second one (E2E-A) includes all three RGB-D data from left, front, and right sensors. This approach aims to demonstrate the potential of utilizing multiple sensors and facilitate a fair comparison between the two models.} Furthermore, we have chosen two versions of the Transfuser method \cite{prakash2021multi}. In the first version called TF-F , the algorithm utilizes a combination of ResNet and transformer \pa{architecture} to process an RGB image and LiDAR data. \pa{In the second version, TF-A, we fed three RGB image and LiDAR data and scale the network to match the input.}
	
	\begin{table*}[t]
		\centering
		\caption{\rpa{Ablation study with task specific metrics}}
		\begin{tabular}{p{2.2cm}p{1.3cm}p{1.6cm}p{1.8cm}p{1.8cm}p{1.6cm}p{1.7cm}p{1.7cm}} 
			\hline
			Model & $\uparrow$ Acc\textsubscript{TL} & $\downarrow$  MAE\textsubscript{SP} & $\downarrow$ BCE\textsubscript{SEG} & $\downarrow$ MAE\textsubscript{WP} &$\downarrow$ MAE\textsubscript{ST} &$\downarrow$ MAE\textsubscript{TH} & $\downarrow$ MAE\textsubscript{BR} \\ 
			\hline
			E2E-F                     &0.9846&NA & 0.1591 & 0.0792  & \underline{0.0174} & 0.0482 & 0.0236   \\
			E2E-A                   &\textbf{0.9882}&NA & 0.0648 & 0.0788  & 0.0195 & \underline{0.0465} & 0.0205 \\
			Ours no SSDC           &0.9812 & 0.2786 & 0.0647 & 0.0827 & 0.0228 & 0.0531 & 0.0332  \\
			Ours no CvT               &0.9871 & \textbf{0.1344} & 0.0634 & \underline{0.0731} & \textbf{0.0173} & 0.0510 &\underline{0.0199}  \\
			Ours no VC                 &0.9839 & 0.3051 & \underline{0.0624} & 0.0817 & 0.0188 & 0.0531 &  0.0257\\
			Ours                      &\underline{0.9878} &\underline{0.2524} & \textbf{0.0620} & \textbf{0.0729} & 0.0182 & \textbf{0.0445} &  \textbf{0.0185} \\ 
			\hline
		\end{tabular}
		\label{table3}
	\end{table*}
	
	\subsection{Results} \label{results}
	In this section, as described in Section \ref{experiments}, we evaluate the proposed model in various scenarios. Table \ref{table1} presents \pa{final} results for our method and the baselines. \rpa{In each column, the optimal value is bolded, and the second-best option is underlined.} Please note that a higher IP or RC does not necessarily indicate better driving performance. A vehicle may complete all routes and receive a high RC, but drive poorly, resulting in low IP and DS, or vice versa. As can be seen from the results of the \pa{Table}~\ref{table1}, our method achieved the highest DS scores for both the 1WN and 1WA scenarios, with RC rates of 99.717 and 91.918, respectively, in Town5 short routes. \rpa{These findings demonstrate both the accuracy (metric RC) and conservativeness (metric DS) of our approach. While E2E-F and E2E-A performed reasonably well, E2E-A showed better performance due to its greater knowledge of the environment. However, TF-F and TF-A both exhibited poor performance in terms of DS and RC. Their IP score is high, since the agent was mostly blocked and did not complete the route. This is likely due to the highly conservative nature of these methods, which cause them to stop frequently during driving, resulting in high IP but low DS.}
	
	\mm{Town5's long routes pose a greater challenge, particularly in adversarial scenarios, where rare accidents or infractions can result in reduced DS. Our method was successful in driving properly, but due to limitations in the training dataset, accidents occurred in some cases. The dataset only collected RGB image data from the top of the expert vehicle, which made it difficult to avoid accidents in situations where the ego vehicle was close to the front vehicle and only the top of the vehicle was visible. As a result, the reported numbers are slightly lower than expected. However, our method achieved better metrics compared to the baselines, demonstrating its great ability. 
		TF-F aimed to drive conservatively, which resulted in high DS due to better IP but lower RC. In contrast, other baselines achieved much higher RC with a lower IP.}  Another crucial factor we focused on is developing an end-to-end algorithm that \mm{is light-weighted and} can be quickly trained with a single advanced GPU. Table~\ref{table2}, illustrates total parameters and GPU memory usage for each baseline as well as our method.
	
	\subsection{Ablation studies on task specific learning}
	\pa{We conducted three ablation studies} to evaluate effectiveness of different modules. The ablations include \mm{removing the side} SDC \mm{map section} (no SSDC), replacing CvT with EfficientNet (no CvT), and removing vehicular controls (no VC). The results are presented in Table~\ref{table3}. Additionally, we analyzed the model's performance in handling multiple perception and control tasks simultaneously by conducting a comparative study with task-specific models to evaluate their intuitive performance on each task independently. \rpa{The metrics scoring include $\text{BCE}_\text{SEG}$ for semantic segmentation, and $\text{Acc}_\text{TL}$ } for traffic light state, similar to \rpa{\citet{natan2022end}}. Mean absolute error is used to justify the model's performance in predicting ego vehicle speed prediction ($\text{MAE}_\text{SP}$), waypoints ($\text{MAE}_\text{WP}$), steering ($\text{MAE}_\text{ST}$), throttle($\text{MAE}_\text{TH}$), and brake ($\text{MAE}_\text{BR}$), which are the same function used for their loss calculation.
	
	\pa{Table \ref{table3}, shows that our approach was more efficient compared to \mmm{E2E}, as it can achieve its minimum only after  21 epochs. Our approach also outperformed \mmm{E2E} in terms of task-specific performance metrics, with more accurate results for vehicular control and waypoint prediction. Although E2E-A had the best traffic light accuracy, our approach provided a more comprehensive solution for autonomous driving.} \pa{The ablation study showed that removing the side maps in SDC decreased the accuracy of SDC, resulting in less accurate vehicular control and waypoint prediction. Conversely, replacing the CvT with Effnet improved the speed, traffic light prediction, and steering control, but reduced the accuracy of other vehicular control commands that require more in-depth knowledge. Finally, removing the vehicular control module improved waypoint predictions but decreased the performance of vehicular control commands. This is due to the lack of an estimator in the system to learn dynamic behavior in the environment, including other vehicles and traffic light state.}
	
	\pa{Furthermore, the results of the ablation study highlight the importance of fusing multiple sensor inputs and the use of deep neural network architectures to achieve better performance in autonomous driving tasks. Future research could explore more efficient architectures for autonomous driving tasks, such as utilizing attention mechanisms instead of concatenation or integrating reinforcement learning algorithms to improve decision-making in dynamic environments. Moreover, the integration of explainable AI techniques would enable us to gain a better understanding of the decision-making process and enhance the transparency and interpretability of autonomous systems.}
	
	\section{Conclusion}
	\label{sec:cn}
	\rpa{In summary, our model stands out as a robust solution for end-to-end autonomous driving, excelling in various scenarios compared to recent models. Utilizing CvT and EfficientNet, our approach effectively addresses challenges in scene understanding and sensor fusion, enabling optimal decision-making in dynamic environments. Notably, the model's adept handling of adversarial scenarios, efficient resource utilization, and enhanced domain knowledge showcase its superiority over existing counterparts. The pivotal role of SDC mapping in achieving robust scene understanding is observed. This feature, combined with RGB features, contributes significantly to the model's ability to obtain valuable insights from the environment. This fusion  ensures that the relationships between them are autonomously learned, preventing the loss of critical information. Two distinct agents within the control module enable our model to generate a wide range of driving options, striking a balance between route completion and incurred infraction penalties. }
	
	\rpa{Prediction of traffic light and ego vehicle speed as separate task, enhances the RGB feature extraction process, contributing to more informed decision-making. Additionally, the use of batch normalization and the inclusion of metadata, such as command, route, and ego vehicle speed, further optimize perception accuracy. The observed efficiency, with fewer trainable parameters and optimized GPU resource utilization, underscores its practical viability. The model's capability to broaden domain knowledge by augmenting sensor inputs with left and right sensors, coupled with its efficient training approach, reflects a holistic understanding of the driving environment. As future direction, one can consider increasing the model capacity and the training time to decrease the infractions while maintaining the high route completion. }
	
	

	{\small
		\bibliographystyle{plainnat}
		\bibliography{egbib}
	}
	
\end{document}